%% file: main.tex
\documentclass[runningheads]{llncs}

\usepackage{iciap}

\usepackage{iciapabbrv}

\usepackage{graphicx}
\graphicspath{{pictures}}
\usepackage{booktabs}

\usepackage{algorithm}
\usepackage{algpseudocode}

\usepackage[accsupp]{axessibility}  

\usepackage{hyperref}
\usepackage{orcidlink}

\begin{document}

\title{When Does Pruning Benefit Vision Representations?} 

\titlerunning{When Does Pruning Benefit Vision Representations?}

\author{Enrico Cassano\inst{1}\orcidlink{0009-0004-6490-6503} \and
Riccardo Renzulli\inst{1}\orcidlink{0000-0003-0532-5966} \and
Andrea Bragagnolo\inst{2}\orcidlink{0000-0002-8619-1586} \and 
Marco Grangetto\inst{1}\orcidlink{0000-0002-2709-7864}}

\authorrunning{E.~Cassano et al.}

\institute{University of Turin, Computer Science Department \\
\email{\{name.surname\}@unito.it} \\
\and LINKS Foundation, AI, Data \& Space (ADS), Torino, IT \\ 
\email{andrea.bragagnolo@linksfoundation.com}}

\maketitle              

\input{sec/0_abstract}    
\input{sec/1_introduction}
\input{sec/2_relatedwork}
\input{sec/3_method}
\input{sec/4_experiments}
\input{sec/5_conclusion}

\bibliographystyle{splncs04}
\bibliography{bibliography/bibliography}


\end{document}

%% file: sec/0_abstract.tex
\begin{abstract}
Pruning is widely used to reduce the complexity of deep learning models, but its effects on interpretability and representation learning remain poorly understood. This paper investigates how pruning influences vision models across three key dimensions: (i) interpretability, (ii) unsupervised object discovery, and (iii) alignment with human perception. We first analyze different vision network architectures to examine how varying sparsity levels affect feature attribution interpretability methods. Additionally, we explore whether pruning promotes more succinct and structured representations, potentially improving unsupervised object discovery by discarding redundant information while preserving essential features. Finally, we assess whether pruning enhances the alignment between model representations and human perception, investigating whether sparser models focus on more discriminative features similarly to humans. Our findings also reveal the presence of ``sweet spots'', where sparse models exhibit higher interpretability, downstream generalization and human alignment. However, these spots highly depend on the network architectures and their size in terms of trainable parameters. Our results suggest a complex interplay between these three dimensions, highlighting the importance of investigating when and how pruning benefits vision representations.

\keywords{Pruning  \and Computer Vision \and Interpretability \and XAI.}
\end{abstract}

%% file: sec/1_introduction.tex
\section{Introduction}

Deep learning models, particularly Convolutional Neural Networks (CNNs) and Vision Transformers (ViTs), have become the state-of-the-art in computer vision, achieving impressive performance across a range of tasks~\cite{rawat2017deep,ye2019cross}. However, the drive towards larger, more complex models has led to architectures with billions of parameters. While these larger models often deliver superior performance, they also come with substantial computational and energy costs, making them difficult to train and deploy efficiently.

To mitigate these issues, \textit{neural network pruning} has emerged as a popular technique to reduce model complexity by removing unnecessary or redundant parameters. This process can significantly lower the computational load and memory requirements, making it more feasible to deploy large models in resource-constrained environments~\cite{rendacomparing,icip, Liao_2023_ICCV}. 

On the other hand, model interpretability is crucial, particularly in high-stakes domains such as healthcare~\cite{wang2023interpretable} and autonomous driving~\cite{jing2022inaction}. Despite their success, deep learning models remain ``black-box'' systems, namely not easily interpretable. This lack of transparency prevents the widespread adoption of AI in applications that demand accountability and explainability.

While significant research has focused on improving the interpretability of unpruned models using techniques such as Integrated Gradients (IG), GradCAM, and Guided GradCAM (G-GradCAM)~\cite{selvaraju2017grad,sundararajan2017axiomatic}, little is known about how pruning impacts a model’s ability to explain its decisions. Since pruning modifies a model’s internal structure by removing less significant parameters, it can either enhance or degrade its interpretability. This brings us to our first research question (\textbf{RQ1}): \textit{Does neural network pruning affect model interpretability?} Beyond interpretability, pruning is known to promote more compact and efficient representations by extracting succinct features. A more compressed model may discard redundant information while preserving the most discriminative features, potentially improving unsupervised object discovery (OD). This raises our second research question (\textbf{RQ2}): \textit{Does pruning enhance object discovery by promoting more concise and structured feature representations?} Furthermore, Sundaram et al.~\cite{sundaram} recently showed that aligning vision models with human perceptual judgments enhances their usability across diverse vision tasks. Since synaptic pruning in humans occurs between infancy and adulthood, by removing neurons that are redundant and strengthening synaptic
connections that are useful~\cite{RAKIC1994227,kolb2009fundamentals,hooker}, it is interesting to explore whether sparser model representations are more aligned with human perception. This leads to our third research question (\textbf{RQ3}): \textit{Do pruned models exhibit better alignment with human perceptual representations?} To address these questions, we conduct extensive experiments on models pre-trained on a source dataset at varying sparsity levels, analyzing how pruning influences interpretability, object discovery, and human alignment (HA).

Our findings can be summarized as follows: (i) interpretability and performance exhibit a negative correlation in CNNs (i.e., higher interpretability $\rightarrow$ lower accuracy), whereas for ViTs the two are more positive correlated (i.e., lower interpretability $\rightarrow$ lower accuracy) (RQ1); (ii) pruning enhances object localization more in CNNs than in ViTs (RQ2); (iii) smaller (e.g., ResNet18, ViT-B/32) architectures better preserve human-alignment than their larger counterparts (RQ3).

%% file: sec/2_relatedwork.tex
\section{Related Work}
\label{sec:sota}
In this section, we review key developments in neural network interpretability and pruning techniques, examining their intersection in recent research. 

\noindent\textbf{Interpreting vision models.} According to the literature, interpretability in deep learning
is ``\textit{the extraction of relevant knowledge from a machine-
learning model concerning relationships either contained
in data or learned by the model}''~\cite{murdoch}. Among the most common vision techniques are IG ~\cite{sundararajan2017axiomatic}, GradCAM~\cite{selvaraju2017grad}, and G-GradCAM~\cite{springenberg2014striving}. IG computes gradients along the path from a baseline to the input, ensuring compliance with sensitivity and implementation invariance axioms. GradCAM generates saliency maps by leveraging gradients from the final convolutional layer to highlight critical image regions, while G-GradCAM enhances precision by integrating GradCAM with Guided Backpropagation.
\noindent\textbf{Model pruning.} It reduces the computational cost of neural networks by removing weights, filters, or neurons, often without significant loss of accuracy, making it beneficial for resource-constrained tasks~\cite{rendacomparing}. The Lottery Ticket Hypothesis even suggests that subnetworks can improve on the original network when trained in isolation~\cite{frankle2018lottery}.
\noindent\textbf{Pruning and interpretability.} 
Abbasi et al.~\cite{abbasi2017interpreting} shows that pruning can identify shape-selective filters as more critical than color-selective ones. Yeom et al.~\cite{yeom2021pruning} prune models based on Layer-wise Relevance Propagation (LRP), improving accuracy when data for transfer is scarce. Despite these interpretability-aware approaches, our work employs a magnitude-based pruning technique, since it's state of the art. Wong et al.~\cite{pmlr-v139-wong21b} showed that sparse linear models over learned dense feature extractors can lead to more debuggable neural networks. In our work, we focus also on sparse backbones, offering a more comprehensive investigation into the effects of sparsity on interpretability. Ghosh and Batmanghelic~\cite{ghosh2023exploring} evaluated pruning's impact on explainability, showing that GradCAM heatmaps of a CNN become inconsistent across pruning iterations, with relevant pixels shifting from objects to background areas as sparsity increases. Von Rad et al.~\cite{von2024investigating} further supported this finding, revealing that there is no significant relationship between interpretability and pruning rate when using mechanistic interpretability score as a measure and GoogLeNet as CNN model. Merkle et al.~\cite{weber2023less} showed that lower compression rates have a positive influence on explainability on the VGG model, while higher compression rates show negative effects. They also identified ``sweet spots'', where both the perceived explainability and the model's performance increases. Our work significantly expands these investigations by evaluating different architectures (convolutional and transformers) and metrics. Furthermore, to the best of our knowledge, this is the first work to comprehensively analyze pruning effects beyond interpretability methods alone, examining how pruning impacts OD capabilities and HA. 

%% file: sec/3_method.tex
\section{Methodology}\label{sec:methodology}
\begin{figure}[h]
    \centering
    \includegraphics[width=.95\textwidth]{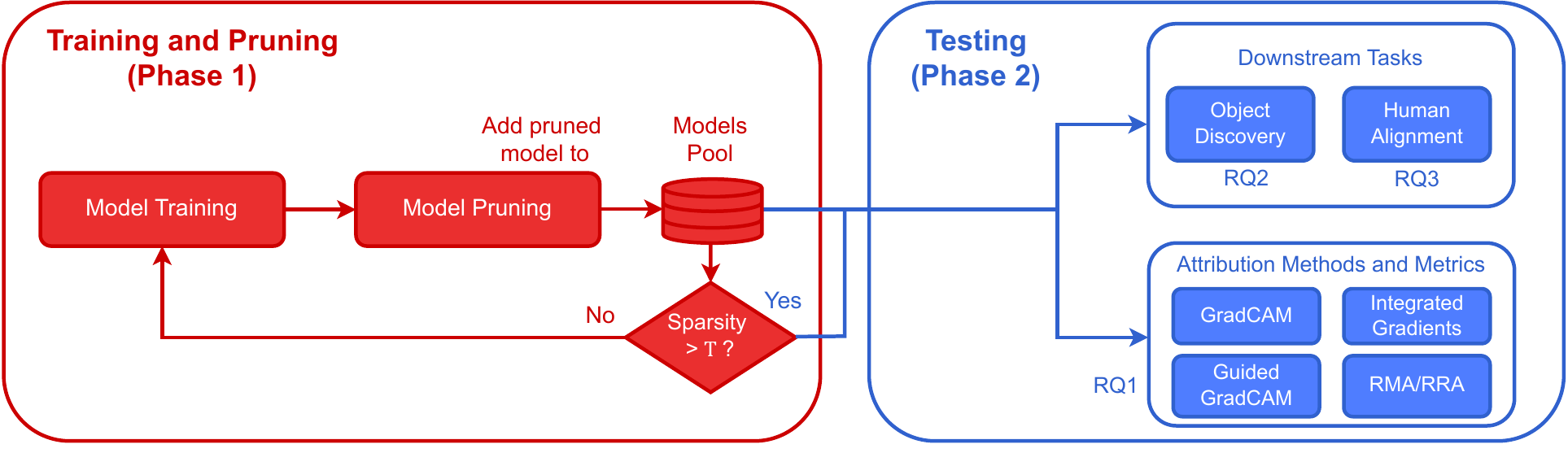}
    \title{Overview of the Experimental Pipeline}
    \caption{We first train and prune a model on a source task at various pruning levels. Then, we assess how pruning affects interpretability (RQ1) and performance on downstream tasks such as OD (RQ2) and HA (RQ3).}
    \label{fig:gp}
\end{figure}

Figure \ref{fig:gp} depicts the general pipeline of our method. 
In the first phase, we train models on a source dataset and progressively apply iterative pruning to increase sparsity until we reach a certain threshold $T$. Pruned models are then pooled for further analysis. In the second phase, we evaluate the interpretability of the pruned models with different attribution methods (GradCAM, IG, G-GradCAM), qualitatively and quantitatively (RQ1). We further benchmark their generalization on downstream tasks such as OD (RQ2) and HA (RQ3).

In the next sections, we detail each step of our pipeline, including the pruning strategy (Section~\ref{sec:pruning-methodology}), interpretability techniques (Section~\ref{sec:attribution}) and downstream tasks(Section~\ref{sec:downstream-tasks}).
\subsection{Pruning}
\label{sec:pruning-methodology}
Our method prunes models with Learning Rate Rewind (LRR), an established variant of Iterative Magnitude Pruning (IMP)~\cite{rendacomparing}, and uses the same original training model recipe. Algorithm~\ref{alg:cap} outlines the main steps of our LRR-modified procedure. 
\begin{algorithm}
\caption{Training and pruning with LRR pseudocode}\label{alg:cap}
\begin{algorithmic}[1]
    \State Initialize an empty pool $P$
    \State Train to completion
    \State Prune the $k$\% lowest-magnitude weights globally
    \State Add the current pruned model with sparsity $S$ to the pool $P$
    \State Retrain using learning rate rewinding with the same training strategy
    \State Return to step 2 until the desired maximum sparsity $T$ is reached.
\end{algorithmic}
\end{algorithm}
LRR continues to train a pruned model from its current state (weights) while repeating the same learning rate schedule in every iteration. Removing the parameter rewinding step has enabled LRR to achieve consistent accuracy gains and improve the movement of parameters away from their initial values~\cite{rendacomparing}. At every pruning iteration, we prune the $k$\% lowest-magnitude weights globally. Then we add each model with a certain sparsity $S$ to our pool, so that we can apply different attribution methods to our pruned models. As described in the next sections, we also evaluate their performance on downstream tasks.  

\subsection{Attribution methods and metrics (RQ1)}
\label{sec:attribution}

We employ several state-of-the-art post-hoc explainability techniques, including IG, GradCAM, and G-GradCAM, to assess model interpretability qualitatively, Relevance Mass Accuracy (RMA) and Relevance Rank Accuracy (RRA) for quantitative evaluation of the attribution maps. 

\noindent\textbf{Qualitative methods.} IG~\cite{sundararajan2017axiomatic} computes attributions by integrating gradients along a linear path from a baseline (e.g., a black image) to the input. This method adheres to the sensitivity and implementation invariance axioms, ensuring reliable attributions. GradCAM~\cite{selvaraju2017grad} creates saliency maps by using gradients from the last convolutional layer to highlight regions in the image important for a specific class prediction. By leveraging the spatial information in convolutional features, GradCAM provides a balance between semantic detail and spatial precision. G-GradCAM further refines GradCAM by combining it with Guided Backpropagation~\cite{springenberg2014striving}, which filters out negative gradient values. This combination yields more focused and precise heatmaps, enhancing model interpretability.

\noindent\textbf{Quantitative methods.}\label{metrics} We employ two common metrics~\cite{arras2022clevr} to evaluate how well a generated saliency map lies in an explanation ground truth segmentation mask ($GT$). These metrics assume that the \textit{majority of the relevance} should be contained within the ground truth mask, either in terms of \textit{relevance mass} or \textit{relevance ranking}. Both metrics produce a score within the range \([0, 1]\), where higher values indicate a more accurate relevance heatmap.
RMA measures the proportion of relevance scores within the ground truth mask relative to the total relevance scores across the image:
    \begin{equation}
    \text{RMA} = \frac{R_{\text{within}}}{R_{\text{total}}}, \quad
    R_{\text{within}} = \sum_{\substack{k=1 \\ p_k \in GT}}^{|GT|} R_{p_k}, \quad
    R_{\text{total}} = \sum_{k=1}^{N} R_{p_k}
\end{equation}
    
    
    
    
    
    where $R_{p_k}$ is the relevance value at pixel $p_k$, $GT$ is the set of pixel
    locations that lie within the ground truth mask, $|GT|$ is the number
    of pixels in this mask and $N$ is the total number of pixels in the image~\cite{arras2022clevr}. So RMA measures how much ``mass'' does the explanation method give to pixels within the ground truth. 
    
RRA measures how many of the highest $K$ relevance scores lie within the ground truth mask.
    Letting $K$ be the size of the ground
    truth mask, we first get the $K$ highest relevance values, then count how many of these values lie within the ground truth pixel locations, and divide
    by the size of the ground truth. Informally, this can be written as
        $P_{top K} = \{p_1, ..., p_K | R_{p_1} > ...> R_{p_K}\}$,
    where $P_{top K}$ is the set of pixels with relevance values $R_{p_1}, ..., R_{p_K}$ sorted
    in decreasing order until the $k$-th pixel.
    Then, the rank accuracy is computed as~\cite{arras2022clevr}:
    \begin{equation}
        \text{RRA} = \frac{|P_{top K} \cap GT |}{|GT|}.
    \end{equation} 

These metrics were computed by evaluating GradCAM, G-GradCAM, IG, and, in the case of transformer-based models, attention maps. 

\subsection{Downstream Tasks}
\label{sec:downstream-tasks}

We also assess how pruning affects model generalization on OD and HA tasks. For OD, we investigate whether pruning can enhance performance by removing noisy or redundant features. For the HA task, it has been observed that neural networks tend to learn high-frequency features (like texture) early in training~\cite{rahaman2019spectral}, while humans are biased towards shapes. Since models (CNNs especially) often rely heavily on texture rather than shape for classification, examining pruned models' robustness to distortions could reveal whether removing parameters pushes models toward using lower-frequency features (like shape). 

\noindent\textbf{Object discovery (RQ2)}. Simeoni et al.~\cite{simeoni2021localizing} proposed LOST, a novel method for unsupervised object localization. The main steps of the technique are: (1) \textit{Initial seed selection.} Seeds are patches that are likely to belong to an object. Patches in an
object correlate positively with each other but negatively with patches in the background. We select patches with the smallest number of positive correlations with other patches as initial seed. 
(2) \textit{Seed expansion.} We expand the initial seed by identifying patches positively correlated with the seed and likely to belong to the same object. 
(3) \textit{Box extraction.} Finally, a binary mask is generated by comparing the features of the seed to all the image features and is used to extract bounding boxes around the detected objects.

We employ as objective score the one introduced by LOST and defined as 
\begin{equation}
    CorLoc = \frac{| LOST \cap GT |}{| LOST \cup GT |}.
\end{equation}

\noindent\textbf{Human alignment (RQ3).} We evaluate our pruned models on the HA task using the established out-of-distribution benchmark by Geirhos et al.~\cite{geirhos2021partial}. This benchmark contains images with various distortions, such as stylization, texture modifications, and synthetic noise. Humans reach near-ceiling accuracies on these challenging test cases as they do on standard noise-free datasets. However, the gap between human and model performance is very large: for example, CNNs trained on ImageNet recognize objects by their textures and not by their shapes, while humans are actually biased towards shapes~\cite{geirhos2018imagenet}. We compute the standard accuracy of our pruned models on these datasets to evaluate if their inductive biases differ from the dense ones.

%% file: sec/4_experiments.tex
\section{Experiments}\label{sec:experiments}

In this section we show how we performed experiments following our evaluation pipeline. First, we describe the datasets, model architectures, and training setup, followed by our results on interpretability analysis and downstream tasks. The source code is publicly available. 
\footnote{\url{https://github.com/EIDOSLAB/pruning-for-vision-representation/}}

\subsection{Setup}

\noindent\textbf{Datasets.}
As our source dataset, we use ImageNet~\cite{ImageNet}, serving as our dataset for model training. We use VOC2012~\cite{Everingham15} for the quantitative interpretability evaluation and the OD task since it contains images and metadata, including the labels of the objects and the bounding boxes of such elements. For assessing HA, we use the 17 datasets introduced in~\cite{geirhos2021partial}, which represent a common benchmark to measure the gap between human and machine vision.

\noindent\textbf{Models architectures and training.}
In our experiments, we employ ViTs and CNNs models. For CNNs, we use ResNet18 and ResNet50, while for ViTs, we employ ViT-B-32 and SwinV2. All models were trained on ImageNet following the training recipe (specific for each model) offered by PyTorch, and, at each iteration, we prune $k=20\%$ of the model's weights until we reach a maximum sparsity threshold $T$ of nearly $100\%$. 

\subsection{Results}
\begin{figure}[h]
    \centering
    \includegraphics[width=0.9\columnwidth]{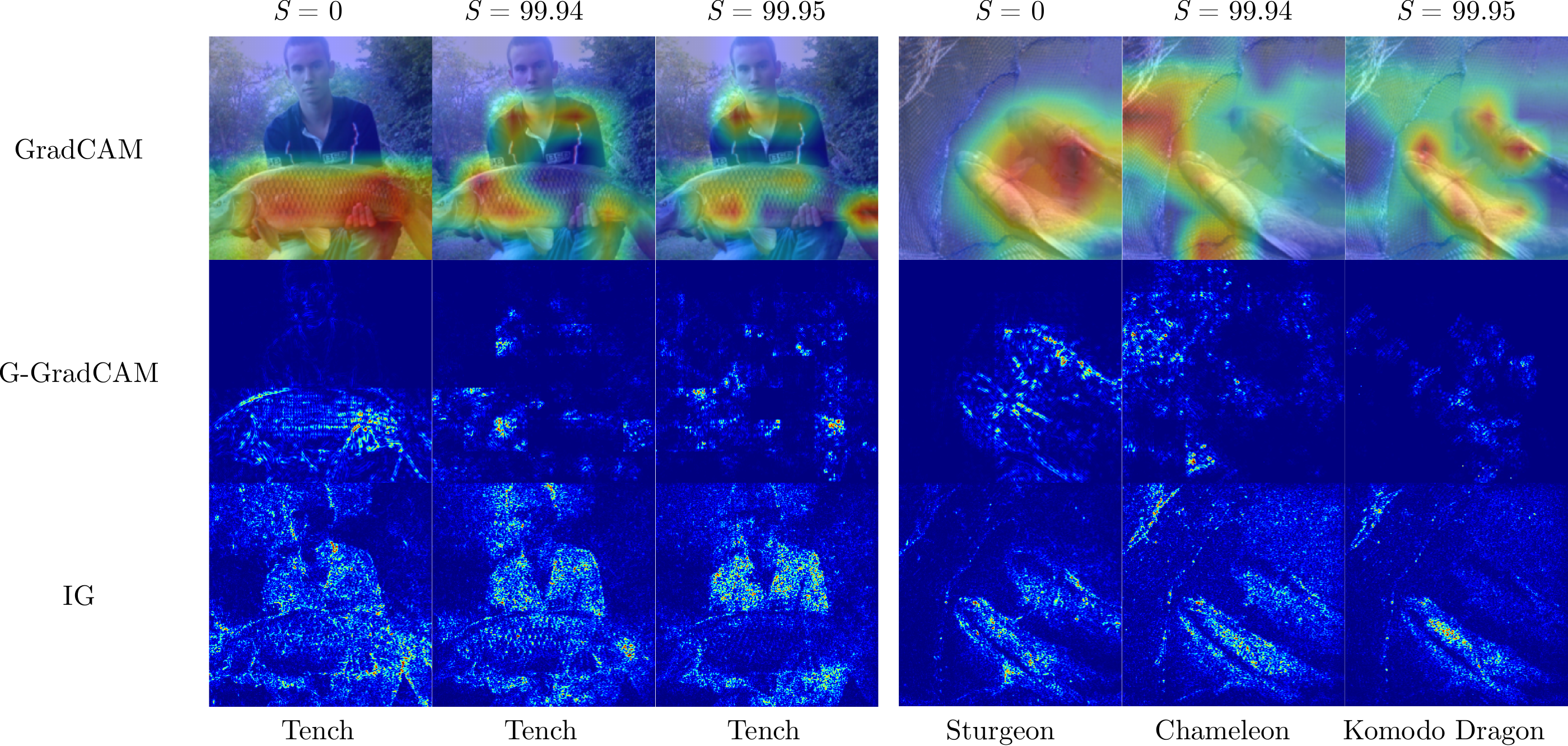}
    \caption{Qualitative evaluation of ResNet50 on two input examples throughout pruning with corresponding model sparsity $S$ and predicted class. 
    }
    \label{fig:qualitative_results}
\end{figure}

\noindent\textbf{Intepretability (RQ1).} 
Figure~\ref{fig:qualitative_results} provides representative examples that illustrate two distinct patterns we observed. 
On the left, we can see a contrastive scenario where attribution quality can degrade even when classification accuracy is maintained. 
On the right, both model predictions and explanations degrade with increased pruning. At $S = 0$, the unpruned ResNet50 predicts a visually similar class (Sturgeon), highlighting semantically meaningful regions, including the fish body. However, as model sparsity increases, the prediction shifts first to Chameleon and then to Komodo Dragon, indicating a progressive semantic drift. Notably, the visualizations begin to emphasize features such as the scale texture and net pattern, elements that may visually resemble reptilian skin, suggesting that these spurious correlations increasingly dominate the model’s focus.

\begin{figure}[!h]
    \centering
    \includegraphics[scale=.18]{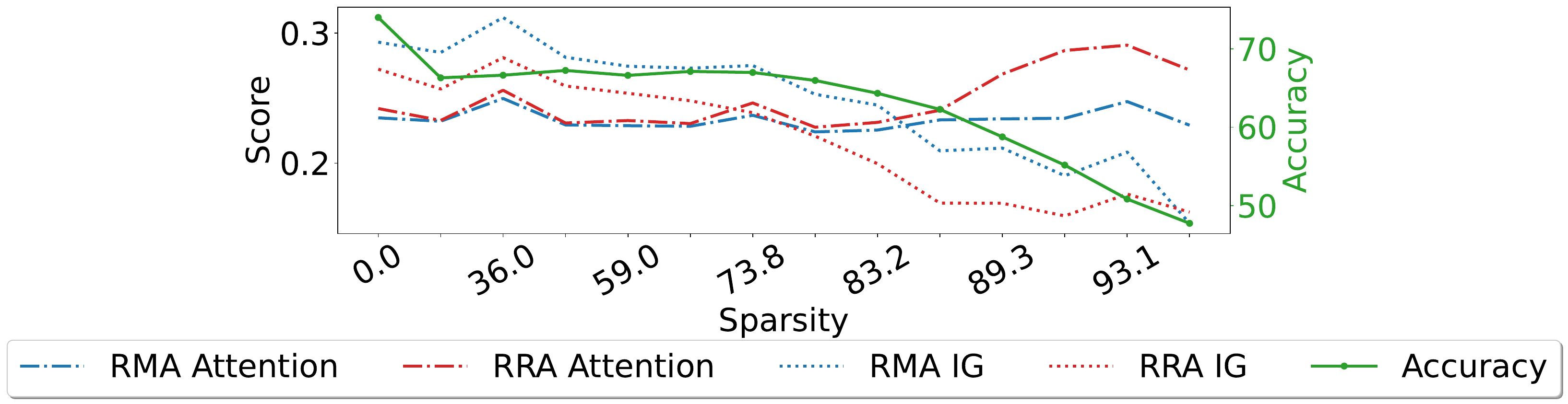}
    \caption{ViT-B-32 RMA and RRA performances.}
    \label{fig:vit-mass_rank}
\end{figure}

For \textit{quantitative} evaluation, the objective metrics employed are RMA and RRA, as described in Section~\ref{metrics}.
\begin{figure}[!h]
    \centering
    \begin{minipage}[b]{0.45\textwidth}
        \centering
        \includegraphics[width=\textwidth]{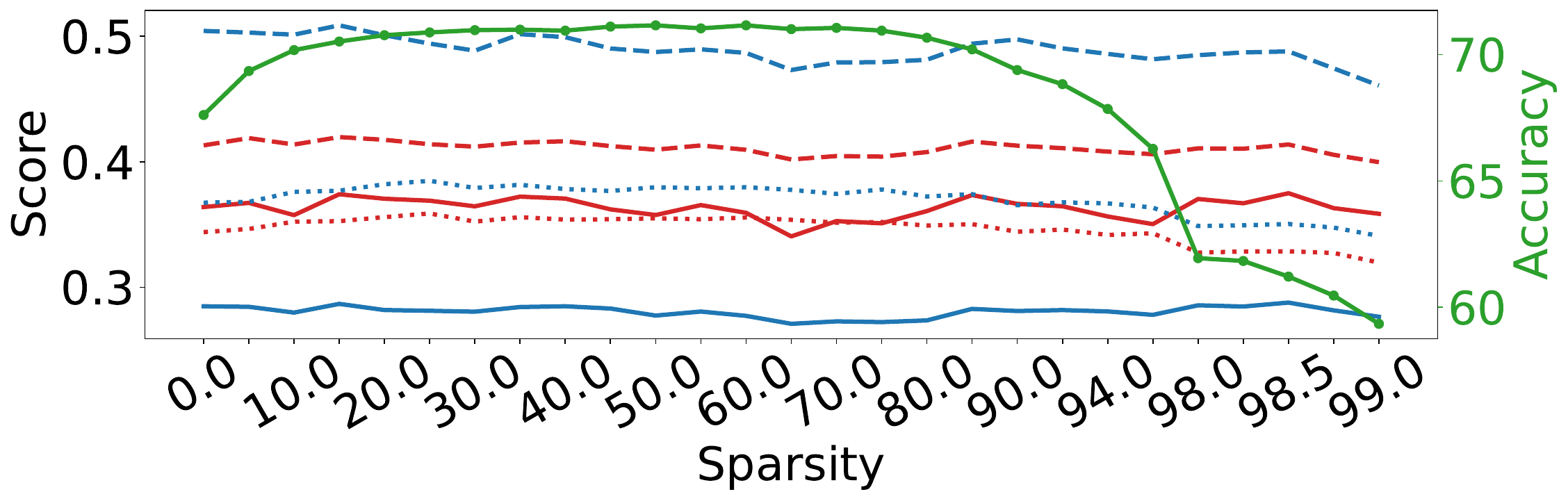}
        \subcaption*{ResNet18}
    \end{minipage}
    \hfill
    \begin{minipage}[b]{0.45\textwidth}
        \centering
        \includegraphics[width=\textwidth]{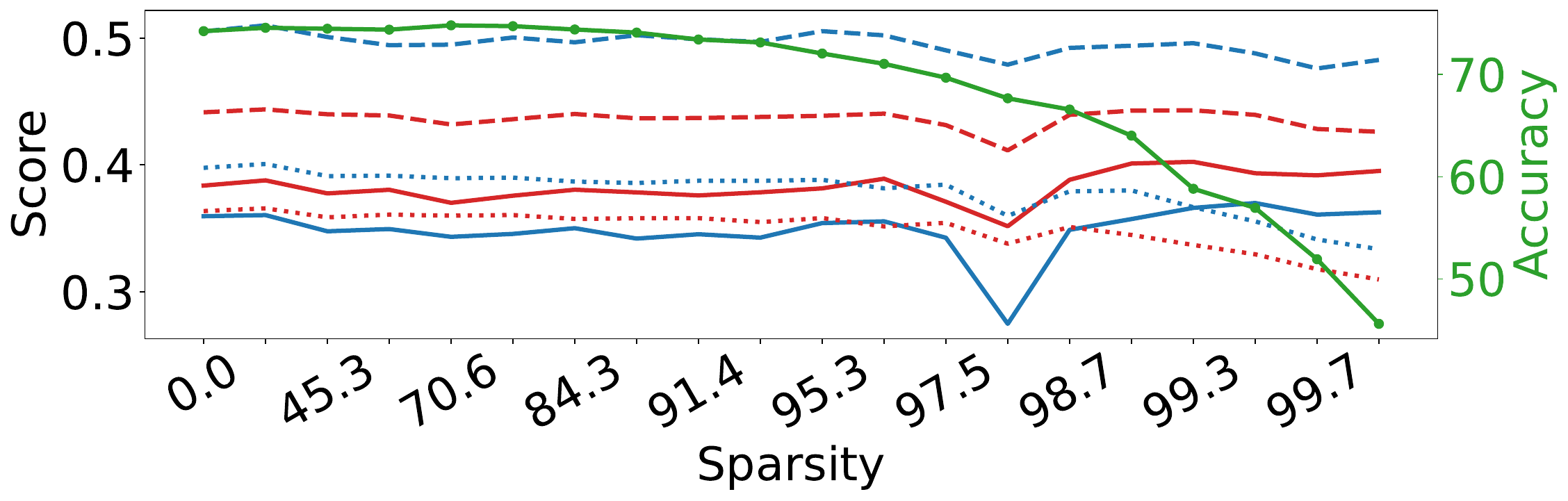}
        \subcaption*{ResNet50}
    \end{minipage}
    
    \vspace{0.5em} 
    
    \begin{minipage}[b]{0.6\textwidth}
        \centering
        \includegraphics[width=\textwidth]{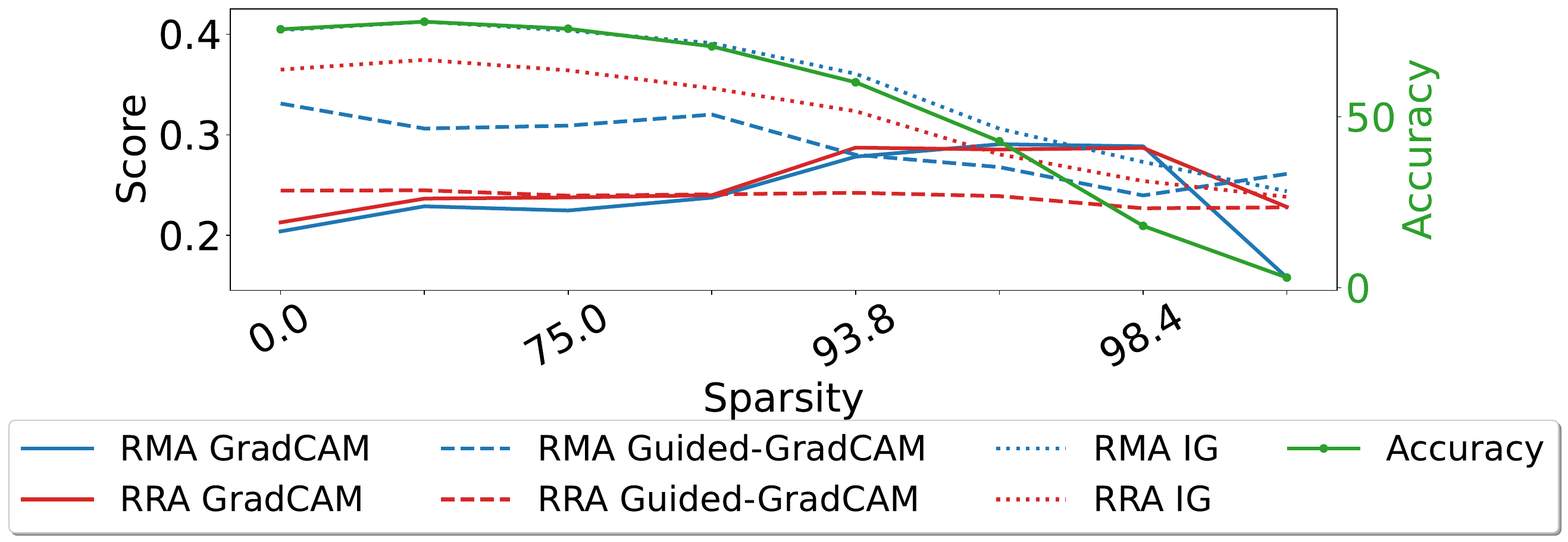}
        \subcaption*{Swin}
    \end{minipage}
    
    \caption{RMA and RRA performances for ResNet18, ResNet50 and Swin.}
    \label{fig:resnet50-mass_rank}
\end{figure}
 We leverage VOC2012 pixel-wise segmentation masks as ground truth for measuring attribution quality.
The evaluation metrics reveal varying and often declining performances across architectures. For CNNs (top-left and top-right plots of Figure \ref{fig:resnet50-mass_rank}), interpretability metrics maintain relatively consistent performances initially, while the accuracy degrades at high sparsity levels. Our findings also confirm the existence of sweet spots~\cite{weber2023less}, i.e. pruning levels where both model accuracy and interpretability metrics simultaneously improve. ResNet18 exhibits a sweet spot at sparsity 15\%, while ResNet50 demonstrates this phenomenon at 25.8\% sparsity. On the other hand, Swin (bottom plot of Figure \ref{fig:resnet50-mass_rank}) shows higher RMA and RRA for GradCAM at high sparsity (between 87.5-98.5\%), but these improvements appear to be specific cases rather than a general trend. Swin shows a partial sweet spot at 50\% sparsity, but RMA on G-GradCAM worsens. ViT-B-32 (Figure \ref{fig:vit-mass_rank}) shows a general deterioration in both RMA and RRA scores for IG as sparsity increases, though attention-based methods remain stable. Unlike the CNNs and Swin, ViT-B-32 does not exhibit any discernible sweet spot. GradCAM and G-GradCAM results were omitted for this model due to poor visualization results.  
Overall, our results reveal a distinct behavior between CNNs and ViTs regarding the relationship between interpretability and performance. For CNNs, interpretability scores (RMA and RRA) on attribution maps (GradCAM, G-GradCAM, and IG) either increase or remain stable as accuracy declines, suggesting that pruning may enhance attributional clarity despite reduced performance (negative correlation, higher interpretability and lower accuracy). 
In contrast, for ViTs, interpretability metrics generally decrease alongside performance (more positive correlation, lower interpretability and lower accuracy), except for RMA and RRA computed on attention maps and GradCAM, where they remain more stable. These findings highlight architectural differences in how sparsity affects model explainability and predictive capability.

\vspace{-3em}
\begin{figure}[!h]
    \centering
    \includegraphics[scale=.18]{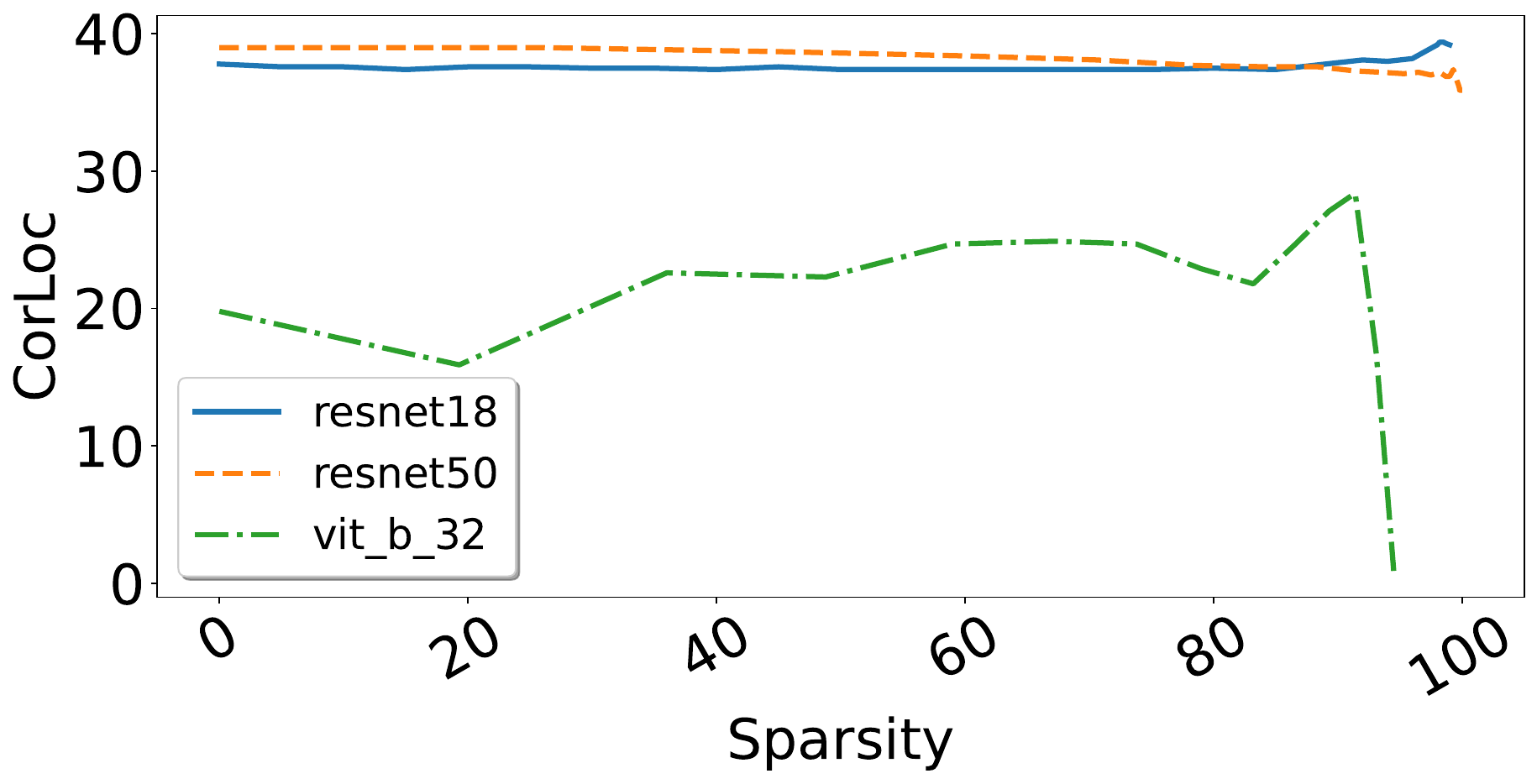}\hspace*{.5cm}
    \caption{ResNet18, ResNet50, ViT-B-32 performances on CORLOC test.}
    \label{fig:global-corloc}
\end{figure}

\noindent\textbf{Object discovery (RQ2).} Figure \ref{fig:global-corloc} shows the CorLoc performance metrics across models on VOC2012. ResNets generally maintain consistent performance across sparsities, with ResNet18 showing a sweet spot at 85\% sparsity where OD performance peaks. Meanwhile, ViT-B-32 exhibits sweet spots at 40\%, 60\% and 90\% sparsity levels, but sharply deteriorates after 95\% sparsity, suggesting a critical threshold beyond which the model's object localization capabilities break down.
Our analysis reveals that sweet spots exist not only for interpretability metrics but also on downstream tasks such as OD.

\begin{figure}[!h]

        \begin{subfigure}[t]{0.49\textwidth}
             \includegraphics[width=\linewidth]{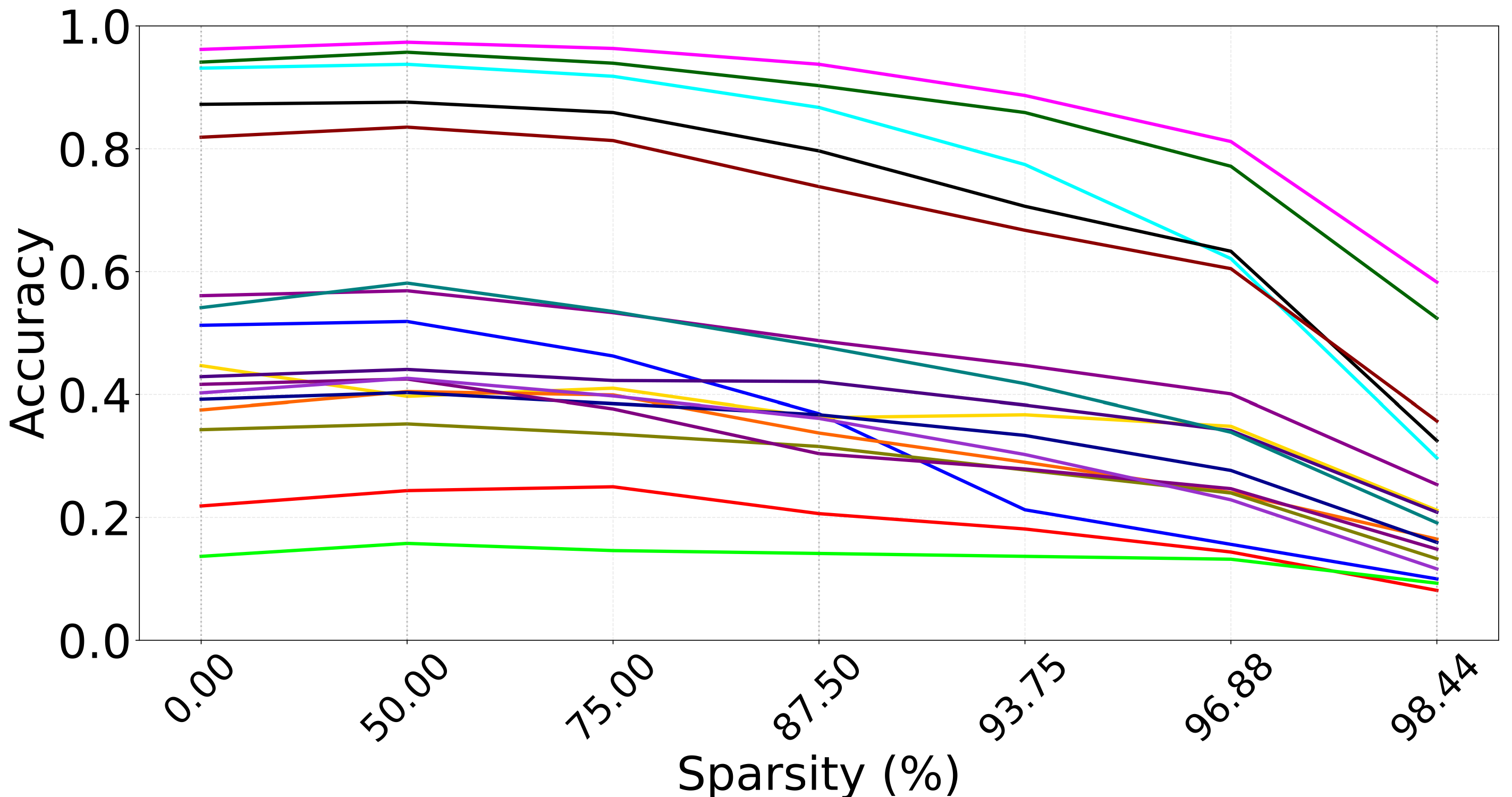}
             \includegraphics[width=\linewidth]{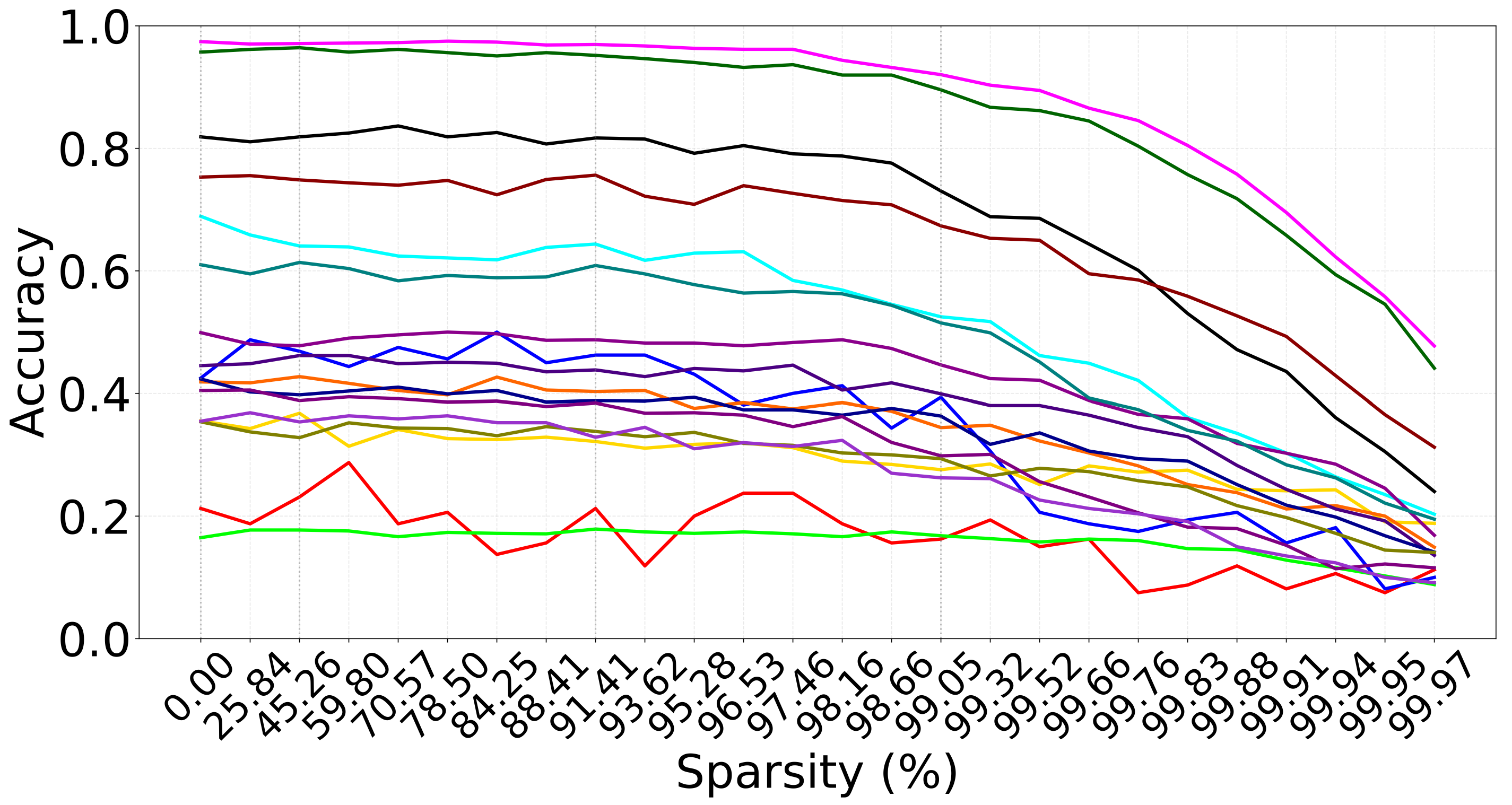}
        \end{subfigure}
        \begin{subfigure}[t]{0.49\textwidth}
             \includegraphics[width=\linewidth]{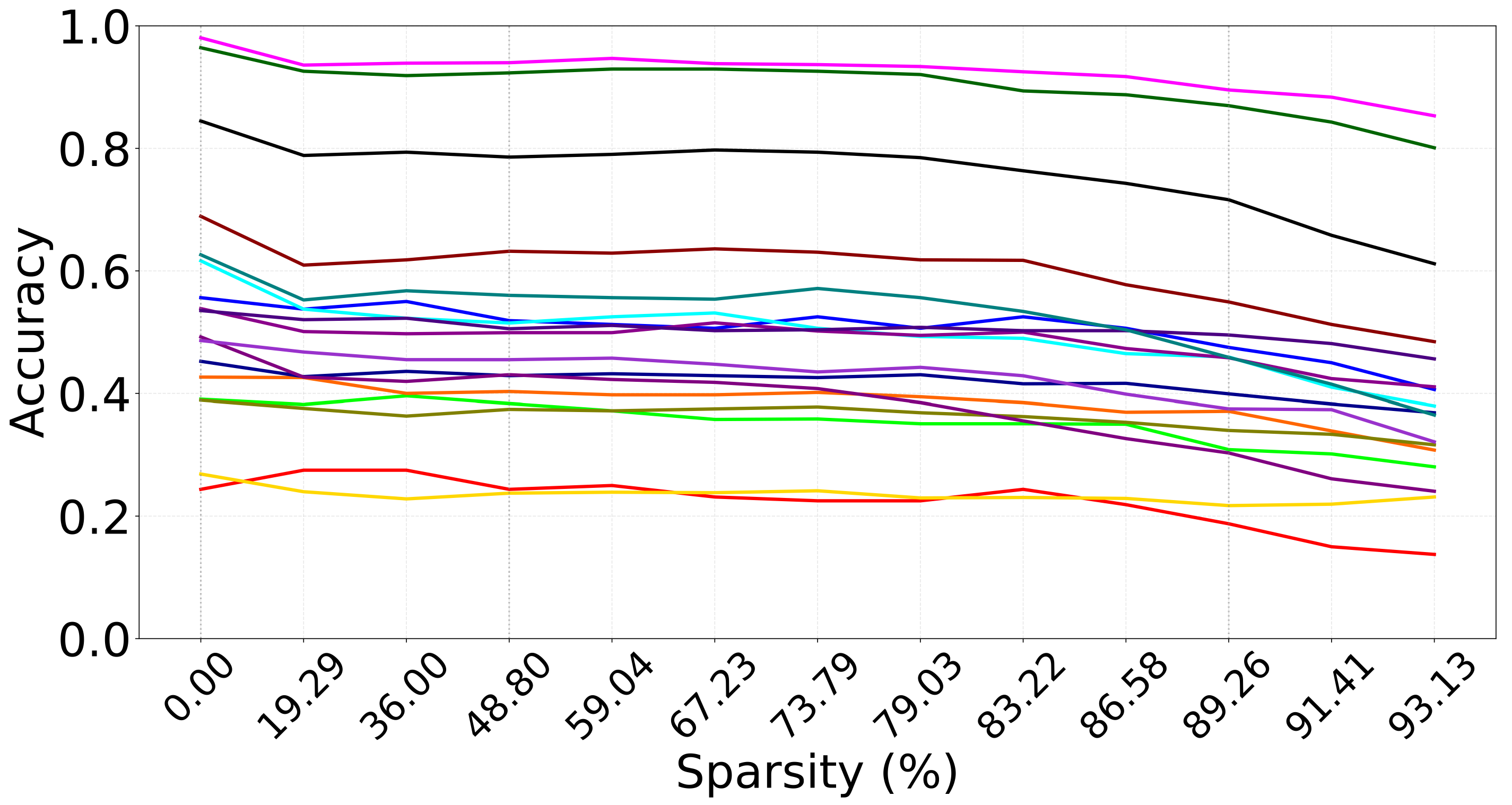}
             \includegraphics[width=\linewidth]{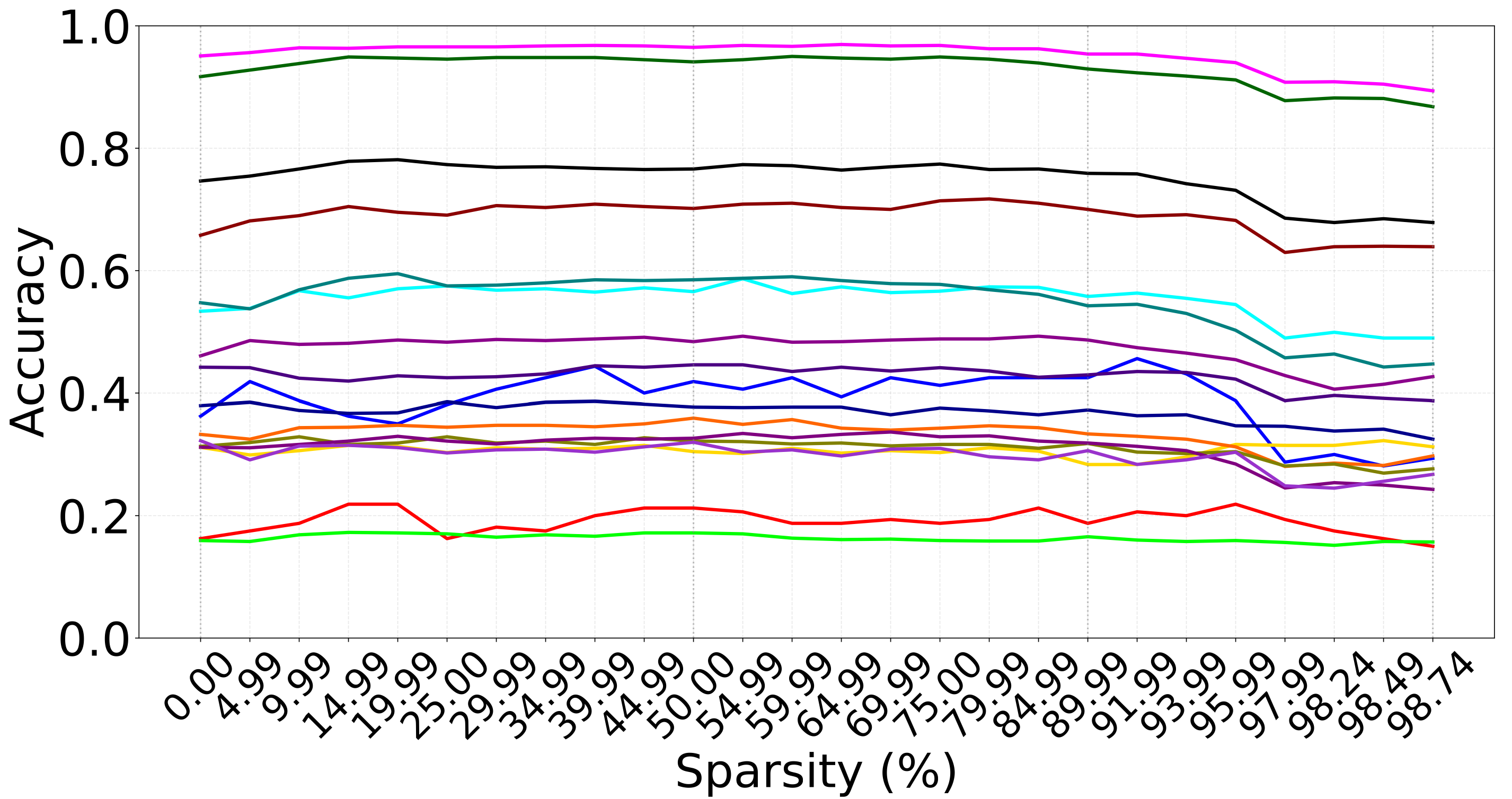}
        \end{subfigure}
        \begin{subfigure}[c]{0.10\textwidth}
             \includegraphics[scale=.145]{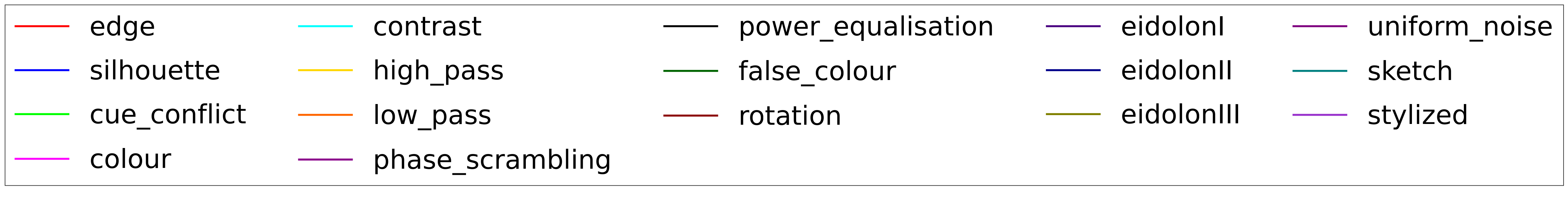}
        \end{subfigure}
    \caption{Models accuracies on the HA task~\cite{geirhos2021partial} (top-left Swin, top-right ViT-B-32, bottom-left ResNet50 and bottom-right ResNet18). }
    \label{fig:mvh-global}
\end{figure}

\noindent\textbf{Human alignment (RQ3).} Figure \ref{fig:mvh-global} reveals the models' behavior across various image distortions. For better visualization, we omitted human performances, but these are near perfect accuracy on each dataset. While the general trend shows performance degradation with increased sparsity, we observe some instances that deviate from such trend. ResNet18's accuracy  (bottom-right plot in Figure \ref{fig:mvh-global}) improves, for example, across silhouette, colour, and false colour datasets. 
We observe sweet spots for HA as well. For example, ResNet18 has multiple of them for silhouette, at pruning steps 1, 8, 10, 12, 14, while on colour and false colour only has one each, on pruning steps 3 and 4 respectively. 
Additionally, Figure \ref{fig:mvh-global} shows how smaller models' performances (top-right: ViT-B-32, bottom-right: ResNet18) degrade less compared to their larger counterparts (top-left: Swin, bottom-left: ResNet50) at high sparsity, showing more consistent alignment towards human perception. 
Figure~\ref{fig:qualitative_results} further demonstrate CNNs' (ResNet50) texture bias, as the model misclassifies images as sturgeon, chameleon, and Komodo dragon instead of tench, all animals sharing a similar scaled skin texture despite having different shapes and contexts.

%% file: sec/5_conclusion.tex
\section{Conclusions}\label{sec:conclusions}

We have shown how pruning affects vision representations across three key dimensions. First, our results revealed that, while pruning generally degrades interpretability, specific configurations show resilience or improvement (RQ1). 
For example, GradCAM on ResNet architectures maintained consistent performances even at high sparsity, while Swin showed improvements between 87.5-98.5\% sparsity. 
Second, we investigated OD capabilities (RQ2), showing that ResNet architectures maintained consistent performance across different pruning levels, while ViT-B-32 performances degrades beyond 95\% sparsity. 
Third, we examined HA (RQ3) across 17 different image distortion datasets. The general trend showed how smaller models are more stable throughout pruning, compared to their larger counterparts.
By identifying effective strategies to enhance model efficiency while maintaining performance across these dimensions, we aim to contribute to the development of more robust and trustworthy AI systems.

\noindent\textbf{Limitations and future work.} Our study opens several insights for future exploration. We focused on a limited set of models, two per architecture class (CNNs and ViTs), and applied a single pruning technique, evaluated using gradient-based explainability methods (IG, GradCAM, and G-GradCAM). 
Although this setup enabled a focused analysis, extending it to a broader range of architectures, pruning strategies, and explainability methods could enhance the generalizability of our findings. Future work could investigate how pruning interacts with other desirable properties such as adversarial robustness and generalization capability. Exploring other compression techniques, including knowledge distillation, quantization, and low-rank decomposition, would further enrich the analysis. Incorporating explainability-aware pruning approaches \cite{yeom2021pruning, abbasi2017interpreting} into the framework could yield informative comparisons to magnitude-based methods.